\documentclass[a4paper,fleqn]{cas-dc}

\usepackage[authoryear]{natbib}
\usepackage[safe]{tipa}
\usepackage{url}
\usepackage{xcolor}
\usepackage{hyperref}
\usepackage{xcolor}
\usepackage{amssymb,amsmath,amsthm}

\usepackage[ruled,noend]{algorithm2e}
\SetKw{KwBy}{by}

\def\tsc#1{\csdef{#1}{\textsc{\lowercase{#1}}\xspace}}
\tsc{WGM}
\tsc{QE}
\tsc{EP}
\tsc{PMS}
\tsc{BEC}
\tsc{DE}
\begin{document}
\let\WriteBookmarks\relax
\def\floatpagepagefraction{1}
\def\textpagefraction{.001}
%\shorttitle{}
%\shortauthors{CV Radhakrishnan et~al.}

\title [mode = title]{A Deep Joint Sparse Non-negative Matrix Factorization Framework for Identifying the Common and Subject-specific Functional Units of Tongue Motion During Speech}                      

\author[1]{Jonghye Woo}\corref{cor1}
\cortext[cor1]{Corresponding author}
%\ead{jwoo@mgh.harvard.edu}
\author[1]{Fangxu Xing}
\author[2]{Jerry L. Prince}
\author[3]{Maureen Stone}
\author[4]{Arnold D. Gomez}
\author[5]{Timothy G. Reese}
\author[5]{Van J. Wedeen}
\author[1]{Georges El Fakhri}

\address[1]{Gordon Center for Medical Imaging, Department of Radiology, Massachusetts General Hospital and Harvard Medical School, Boston, MA 02114, USA}
\address[2]{Department of Electrical and Computer Engineering, Johns Hopkins University, Baltimore, MD 21218, USA}
\address[3]{Department of Neural and Pain Sciences, University of Maryland School of Dentistry, Baltimore, MD 21201, USA}
\address[4]{Department of Neurology, Johns Hopkins University School of Medicine, Baltimore, MD 21218, USA}
\address[5]{Athinoula A. Martinos Center for Biomedical Imaging, Department of Radiology, Massachusetts General Hospital and Harvard Medical School, Boston, MA 02129, USA}

\begin{abstract}
Intelligible speech is produced by creating varying internal local muscle groupings---i.e., functional units---that are generated in a systematic and coordinated manner. There are two major challenges in characterizing and analyzing functional units.~First, due to the complex and convoluted nature of tongue structure and function, it is of great importance to develop a method that can accurately decode complex muscle coordination patterns during speech. Second, it is challenging to keep identified functional units across subjects comparable due to their substantial variability. In this work, to address these challenges, we develop a new deep learning framework to identify common and subject-specific functional units of tongue motion during speech.~Our framework hinges on joint deep graph-regularized sparse non-negative matrix factorization (NMF) using motion quantities derived from displacements by tagged Magnetic Resonance Imaging. More specifically, we transform NMF with sparse and graph regularizations into modular architectures akin to deep neural networks by means of unfolding the Iterative Shrinkage-Thresholding Algorithm to learn interpretable building blocks and associated weighting map. We then apply spectral clustering to common and subject-specific weighting maps from which we jointly determine the common and subject-specific functional units. Experiments carried out with simulated datasets show that the proposed method achieved on par or better clustering performance over the comparison methods.Experiments carried out with \textit{in vivo} tongue motion data show that the proposed method can determine the common and subject-specific functional units with increased interpretability and decreased size variability.

\end{abstract}


\maketitle

%% main text
\section{Introduction}
\label{intro}

Intelligible speech is produced by intricate and successful orchestration of local muscle groupings---i.e., functional units---of the extremely complex muscular architecture of the tongue~\citep{woo2018sparse}. The tongue is an organ that is controlled intricately by the support of its myoarchitecture, comprising an array of highly inter-digitated intrinsic and extrinsic muscles~\citep{gaige2007three}. As a result, it is of great interest and need to study the intrinsic dimension-reduced structures of speech movements in order to better understand the mechanisms by which intrinsic and extrinsic muscles of the tongue coordinate to generate rapid yet accurate speech movements.~To date, a great deal of work from different disciplines, including neurophysiology, biomechanics, speech and language, and medical imaging and analysis, has hypothesized and demonstrated that the control of tongue movements is governed by a reduced number of degrees of freedom~\citep{gick2013modularizing} that is associated with corresponding neuromuscular modules~\citep{bizzi1991computations, kelso2009synergies}, or fixed or mutable local muscle groupings~\citep{woo2018sparse,stone2004functional}. 

%%%%%%%%%%%%%%%%%%%%%%%%%%%%%%%%%%%%%%%%%%%%%%%%%%%%%%%%
\def\FigureHeight{72mm}
\begin{figure*}[t]
 \center{
 \begin{tabular}{c@{ }c}
   \includegraphics[trim=0mm 0mm 0mm
0mm,clip=true,height=\FigureHeight]{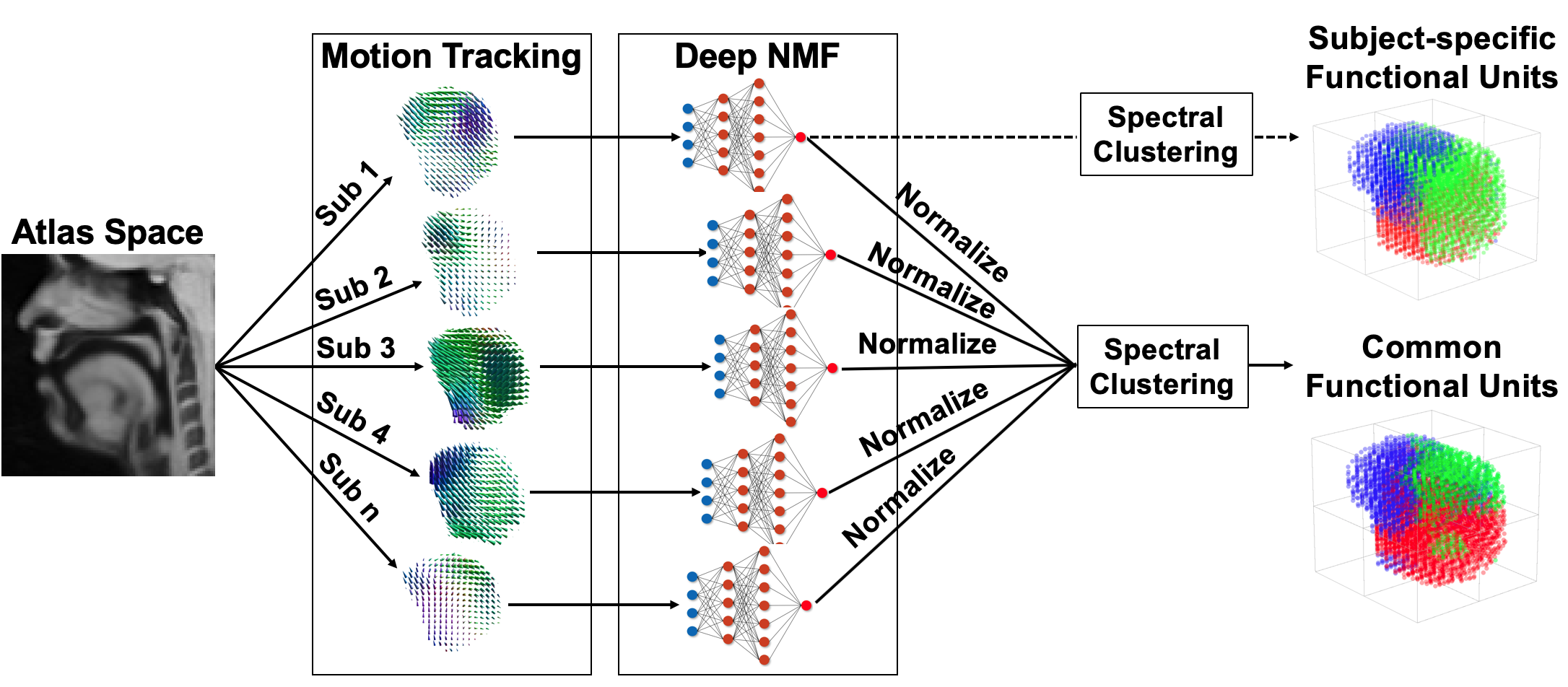}
& \hspace{0pt} 
\end{tabular}}
\caption{A flowchart of our method. Subject-specific motion tracking results from tagged MRI are first transformed into an atlas space representing a neutral tongue position. Our deep learning framework is then used to determine the common as well as subject-specific functional units.}\label{fig:flowc} 
\end{figure*}

Medical imaging techniques, including magnetic resonance imaging (MRI), have been used to characterize functional units of speech movements~\citep{woo2018sparse, stone2004functional}. In particular, tagged MRI allows us to non-invasively track spatiotemporally varying speech movements at the voxel level~\citep{parthasarathy2007measuring, xing2017phase, osman1999cardiac}.~More specifically, MR tagging can generate temporary grid-like patterns via a sequence of radiofrequency pulses within the tissue. This is achieved by spatially modulating longitudinal magnetization of hydrogen protons. As a result, the induced temporary grid-like tagging patterns deform alongside tongue motion and are visible perpendicular to tagging planes. Then, 2D plus time or 3D plus time velocity fields at the voxel level are typically estimated via tracking algorithms based on harmonic phase (HARP)~\citep{parthasarathy2007measuring, osman1999cardiac, xing2017phase}.

In order to identify the ``manageable'' number of degrees of freedom of speech movements and better understand their spatial and temporal couplings between different parts of the tongue, various modeling approaches have been developed. Non-negative matrix factorization (NMF)~\citep{lee1999learning} and its variants including sparse NMF are well-recognized, given that NMF is capable of examining signals derived from intrinsic muscle activations that are non-negative ~\citep{ting2010decomposing}. Sparse NMF~\citep{kim2008sparse} is a matrix decomposition approach, where an input matrix whose entries are non-negative is expressed as a sparse linear combination of a set of building blocks. Since building blocks can be seen as an underlying anatomical basis, its associated weighting map can be used to reveal consistent and coherent sub-motion patterns. To further characterize the underlying physiology of speech movements using NMF, additional prior knowledge, such as manifold geometry of input movement data, has also been investigated~\citep{woo2018sparse, cai2010graph}.

There are two major challenges to be addressed in this work. First, the prior approach~\citep{woo2018sparse} to identify functional units using sparse NMF is based on a shallow NMF model, which may not capture the underlying tongue's complex physiology accurately. The production of speech requires the complexity inherent in the execution, involving the activation of thousands of motor units in orthogonally oriented and interdigitated muscles. In addition, functional units are seen as 3D localized regions that show coherent displacement or related quantities, which are intermediate structures that interface between tongue muscle activation and tongue surface motion~\citep{woo2018sparse}. Accordingly, there is a need to develop an NMF model that can learn complex muscle coordination patterns from motion features derived from speech movements, while retaining the constraints and advantages of an NMF model which can deal with non-negative signals and offer parts-based and interpretable representations, respectively. In addition, a deep NMF is required, which can interrogate the relationship between complex muscle interdigitation and local activation and tongue surface motion~\citep{woo2015high, stone2018structure}. Second, because of the different motions that tongues produce during the course of speech, functional units vary substantially from one subject to another. Thus, one of the important hurdles in analyzing functional units is how to keep identified functional units across subjects comparable due to their substantial variability. Independently applying an NMF model to determining individual functional units may result in identifying building blocks and their weighting that are suboptimal, thereby yielding results that are challenging to objectively compare across subjects.

To alleviate the aforementioned challenges, we present a normalization method that can identify both the common and subject-specific functional units in a cohort of speakers in an atlas space from tagged MRI and 3D plus time voxel-level tracking by extending our prior work~\citep{woo2020identifying}. In contrast to the prior work~\citep{woo2020identifying}, we further describe a refined method using a deep joint sparse NMF framework to identify spatiotemporally varying functional units using a simple utterance and carry out extensive validations on both simulated and \textit{in vivo} tongue motion data. Our deep joint sparse NMF framework computes a set of building blocks and both subject-specific and common weighting maps given motion quantities from tagged MRI. We then apply spectral clustering to the common and subject-specific weighting maps to jointly determine the common functional units across subjects and the subject-specific functional units for each subject.

In addition, we incorporate both sparse and graph regularizations into our framework. First, we impose a sparsity constraint on the weighting map to obtain the optimized and simplest functional units of tongue motion, which is consistent with the notion of ``gestures” in phonological theories~\citep{ramanarayanan2013spatio}. Second, we impose a graph regularization, which allows us to discover the intrinsic geometric structure of the motion data. Since a Euclidean distance measure via Frobenius norm is used in this work, the intrinsic and manifold geometry of the input motion data is largely ignored. By incorporating both regularizations into our formulation, we can determine a set of simplest and intrinsic sub-motion patterns by promoting the computation of distances on a manifold. As well, it is possible to identify a low-dimensional yet interpretable subspace from tongue motion data.

The contributions of the proposed method can be summarized as follows:
\begin{itemize}
  \item The most prominent contribution of this work is to construct an atlas of the functional units---i.e., the common consensus functional units---of how tongue muscles coordinate to produce target observed motion in a healthy population from cine and tagged MRI.
  \item This proposed work can simultaneously yield both common as well as subject-specific functional units within a material coordinate system with reduced size variability, thereby greatly facilitating the comparison of identified functional units during speech across subjects.
  \item This proposed work converts NMF with sparse and graph regularizations into modular architectures by means of unfolding Iterative Shrinkage-Thresholding Algorithm (ISTA), thereby accurately capturing the sub-motion patterns through each subject's underlying low-dimensional subspace.
  \item This proposed work achieves on par or better clustering performance over the comparison methods on both simulated and \textit{in vivo} tongue motion datasets.  
  \item Experiments carried out with \textit{in vivo} tongue motion data show that the proposed method can determine the common and subject-specific functional units with increased interpretability and decreased size variability.
\end{itemize}

The rest of this paper is structured as follows. Section~\ref{sec:related_work} reviews related work. Section~\ref{sec:proposed} defines the problem and describes our proposed approach. The experimental results are shown in Section~\ref{sec:result}, and Section~\ref{sec:discussion} presents a discussion. Finally, we conclude this paper in Section~\ref{sec:conclusion}.

\section{Related Work}\label{sec:related_work}
\subsection{Functional Units}
Various attempts have been made to investigate functional units of tongue motion during speech using imaging and motion capture techniques. For example, \cite{green2003tongue} studied functionally independent articulators within the tongue based on a correlation analysis from an x-ray microbeam database. In that work, the functional independence was assessed through movement coupling relations, demonstrating phonemic differentiation in vertical tongue motions from 20 vowel-consonant-vowel (VCV) combinations. Similarly, \cite{stone2004functional} examined the functional independence of five segments within the tongue during speech using a correlation analysis from 2D plus time ultrasound and tagged MRI. That work demonstrated that adjacent segments have high correlations, while distant segments have negative correlations consistent with linguistic constraints. The present work improves upon the previous work~\citep{stone2004functional} by incorporating 3D plus time tagged MRI to assess how different parts of the tongue coordinate during speech. \cite{ramanarayanan2013spatio} proposed a computational framework to identify linguistically interpretable tongue movement primitives of speech articulation data based on a convolutive NMF algorithm with sparsity constraints from electromagnetic articulography and synthetic data generated via an articulatory synthesizer. Our proposed work is inspired by this approach~\citep{ramanarayanan2013spatio}, but we use far richer 3D plus time displacements from tagged MRI together with deep NMF with the addition of sparsity and intrinsic data geometry in identifying functional units. \cite{woo2018sparse} presented a framework to examine functional units using a shallow graph-regularized sparse NMF model from tagged MRI and 3D plus time voxel-level tracking. Recently, \cite{sorensen2019task} investigated a functional grouping of articulators and its variability across participants from real-time MRI. All of this work, however, studied subject-specific functional units and therefore lacked an understanding of the common functional units in a healthy population. In this work, we extend our prior approaches~\citep{woo2018sparse, woo2020identifying} to develop a deep joint sparse  NMF framework that can co-identify common and subject-specific functional units across participants.

\subsection{Deep NMF}

The recent success of deep neural networks allowed many researchers to investigate ``deep NMF.'' For example, a deep unfolding method was developed, yielding a new formulation that can be trained using a multiplicative back-propagation method~\citep{hershey2014deep}. In addition, deep NMF~\citep{le2015deep} was proposed by unfolding the NMF iterations and untying its parameters for the application of audio source separation. Furthermore, a new architecture combining NMF with deep recurrent neural networks~\citep{wisdom2017deep} was presented by unfolding the iterations of ISTA~\citep{gregor2010learning}. In the present work, we aim to develop deep NMF with both sparse and graph regularizations by unfolding the iteration of ISTA. We note that a similar idea has been explored in the prior work~\citep{hershey2014deep,le2015deep,wisdom2017deep} described above, but this work further incorporates both sparse and graph regularizations into the deep NMF framework.

\section{Methodology}\label{sec:proposed}
\subsection{Participants and MRI Data Collection}

In this work, a total of 18 healthy speakers were included. Table~\ref{table:subject} lists the characteristics of subjects. All subjects are native speakers of American English with a Maryland accent. Each speaker was trained before the MR scan to speak a simple utterance (``a souk'') in line with a periodic metronome-like sound. This word is one of several that were chosen by design to move the tongue in specific directions, while being short enough to be spoken during the 1-second recording limit imposed by tag fading. The task begins with the /\textschwa/, which positions the tongue such that the vocal tract tube has an almost uniform cross-sectional area throughout its length. The tongue moves to an anterior position for /s/ and then posteriorly into /u/ and /k/. The vowel /u/ uses a closed jaw, as do the consonants, requiring that all vocal tract shaping be done by deforming the tongue, not merely opening and closing the jaw as can happen during /\textschwa/. Thus, the word keeps the tongue high, maximizes tongue deformation, and moves posteriorly primarily.

Each speaker repeated the speech word following the periodic sound, while acquiring T2-weighted 2D tagged and cine MRI through a Siemens 3.0 T Tim Trio system (Siemens Medical Solutions, Erlangen, Germany) with a 12-channel head coil and 4-channel neck coil. Both dynamic MR images were acquired at 26 frames per second with three orthogonal directions, including coronal, axial, and sagittal directions. Then, for cine MRI, a super-resolution volume reconstruction technique~\citep{woo2012reconstruction} was used to combine three orthogonal stacks to yield a single volume with isotropic resolution. 

%% Table 1 %%%%%%%%%%%%%%%%%%%%%%%%%%%%%%%%%%%%%%%%%%%%%%%%%%%%%%%
\begin{table}[h]
\caption{Characteristics of 18 healthy subjects}
\centering
%\scriptsize{ 
\begin{tabular}{c|c|c||c|c|c} \hline
 \hline 
  Subject & Age & Gender & Subject & Age & Gender \\
   \hline \hline
  1 & 23 & M & 10 & 26 & F \\
 \hline 
  2 & 31 & F & 11 & 22 & M \\
 \hline
  3 & 27 & F & 12 & 43 & M \\
 \hline
  4 & 41 & F & 13 & 27 & M \\
 \hline
  5 & 35 & M & 14 & 42 & F \\
 \hline
  6 & 45 & F & 15 & 59 & F \\
 \hline
  7 & 27 & F & 16 & 52 & M \\
 \hline
  8 & 22 & F & 17 & 54 & M \\
 \hline
  9 & 22 & F & 18 & 27 & M \\
 \hline

\end{tabular}\label{table:subject}
\end{table}
%%%%%%%%%%%%%%%%%

\subsection{Estimation of Subject-specific Motion Fields from Tagged MRI}
For the 3D plus time motion estimation, we use a tracking method by \cite{xing2017phase} that hinges on symmetric and diffeomorphic registration with HARP phase volumes to yield a sequence of voxel-level motion fields during the speech tasks from tagged MRI. In brief, 2D slices into 3D voxel locations are interpolated using cubic B-spline. Then, a HARP tracking method~\citep{osman1999cardiac} is utilized to yield HARP phase volumes. Finally, the iLogDemons method~\citep{mansi2011ilogdemons} is applied to finding symmetric and diffeomorphic transformations from a reference time frame to the target time frame. The transformations are given by 

\begin{equation}\label{transform}
\varphi_{i,j}: \mathrm{\Omega} \to \mathrm{\Omega}, i = 1,\cdots, N,
\end{equation}
where $N$ denotes the number of subjects (in this work, $N$=18), and $j$ denotes the time frame index (i.e., $j$ = 1,$\cdots$, $M$), where $M$ denotes the total number of time frames for the utterance (in this work, $M$=26) in these phase volumes. Finding symmetric and diffeomorphic transformations with the volume-preserving constraint is crucial for tongue motion analysis because the tongue's volume remains invariant, and we need to preserve the smoothness of anatomical details within the tongue in the course of transformation.

\subsection{Identification of Subject-specific Functional Units via a Deep Sparse NMF Framework}

Assume that the tongue is comprised of $K$ distinct clusters---i.e., functional units---in the course of a given phoneme of interest, each of which corresponds to a characteristic motion from which muscle coordinations and interactions occur. In this work, we opt to use graph-regularized sparse NMF to identify functional units for the following reasons. First, in order to accurately characterize each functional unit, it is necessary to project the high-dimensional and complex 3D plus time voxel-level tracking into a $K$ low-dimensional subspace in which each axis corresponds to a particular sub-motion pattern. In addition, it is natural that functional units comprising a subset of intrinsic and extrinsic muscles are not entirely independent of each other; and there could be some overlaps among them. Furthermore, since it is assumed that functional units are the result of an additive mixture of the underlying muscle activations, the linear combination coefficients---i.e., weighting maps---need to take non-negative values only. 

\subsubsection{Extraction of Motion Features}
We extract motion quantities from the 3D plus time motion estimation stated above to identify the functional units~\citep{woo2018sparse}: the magnitude and angle of each trajectory given by
\begin{equation}
\mathcal{M}^{p}_{l} = \sqrt{(x^{p}_{l+1}-x^{p}_{l})^2 + (y^{p}_{l+1}-y^{p}_{l})^2 + (z^{p}_{l+1}-z^{p}_{l})^2}
\end{equation}
\begin{equation}
\mathcal{Z}_l^p  = \frac{{x_{l+1}^p  - x_l^p }}
{{\sqrt {(x_{l + 1}^p  - x_l^p )^2  + (y_{l + 1}^p  - y_l^p )^2 } }} + 1
\end{equation}
\begin{equation}
\mathcal{X}_l^p  = \frac{{y_{l+1}^p  - y_l^p }}
{{\sqrt {(y_{l+1}^p  - y_l^p )^2  + (z_{l+1}^p-z_l^p )^2 } }}+ 1
\end{equation}
\begin{equation}
\mathcal{Y}_l^p  = \frac{{z_{l+1}^p - z_l^p }}
{{\sqrt {(z_{l+1}^p  - z_l^p )^2  + (x_{l+1}^p-x_l^p )^2 } }}+ 1,
\end{equation}
where $\mathcal{M}^{p}_{l}$ is the magnitude of each point trajectory and $\mathcal{Z}_l^p$, $\mathcal{X}_l^p$, and $\mathcal{Y}_l^p$ represent the cosine of the angle after projecting two consecutive adjacent point trajectories in the $z$, $x$, and $y$ axes plus one to make sure that all values are non-negative, respectively.

We then combine all the motion features into a single $4(L-1)\times P$ non-negative matrix $\mathbf{U} =
[\mathbf{u}_1,...,\mathbf{u}_n] \in \mathbb{R}^{m\times n}_+$, where the $p$-th column is expressed as 

\begin{equation}\label{eq:feature}
\mathbf{u}_p = [\mathcal{M}_1^p \cdots \mathcal{M}_{L-1}^p \;\; \mathcal{Z}_1^p \cdots \mathcal{Z}_{L-1}^p \;\; \mathcal{X}_1^p \cdots \mathcal{X}_{L-1}^p \mathcal{Y}_1^p \cdots \mathcal{Y}_{L-1}^p ]^T.
\end{equation}

\subsubsection{Deep Graph-regularized Sparse NMF}

Mathematically, the objective of NMF is to factorize the non-negative matrix $\mathbf{U} = [\mathbf{u}_1,...,\mathbf{u}_n] \in \mathbb{R}^{m\times n}_+$ into the non-negative matrix $\mathbf{V} = [v_{ik}] \in \mathbb{R}^{m\times k}_+$, the building blocks, and the non-negative matrix $\mathbf{W} = [w_{kj}] \in \mathbb{R}^{k\times n}_+$, the weighting map, that minimizes the following objective function:

\begin{equation}\label{eq:nmf}
\mathcal{E}_1 =  \frac{1}{2}\left\| {\mathbf{U} - \mathbf{VW}} \right\|_F^2,
\end{equation}
where $\left\|  \cdot  \right\|_F$ represents the matrix Frobenius norm defined as

\begin{equation}
\left\| \mathbf{V} \right\|_F = \sqrt{\mathrm{Tr}(\mathbf{VV}^T)} = \sqrt{\sum\limits_{i = 1}^m{\sum\limits_{j = 1}^n {v_{ij}^2}}}. 
\end{equation}
Here, $\mathrm{Tr}(\cdot)$ denotes the trace of a matrix. Among other divergence measures~\citep{lee1999learning, cichocki2008non, sra2006generalized}, we focus on the Frobenius norm to compute dissimilarity between the non-negative input data matrix $\mathbf{U}$ and its approximation $\mathbf{VW}$. The objective function of graph-regularized sparse NMF can be defined as: 

%%%%%%%%%%%%%%%%%%%%%%%%%%%%%%%%%%%%%%%%%%%%%%%%%%%%%%%%
\def\FigureHeight{18mm}
\begin{figure}[t]
 \center{
 \begin{tabular}{c@{ }c}
   \includegraphics[trim=0mm 0mm 0mm
0mm,clip=true,height=\FigureHeight]{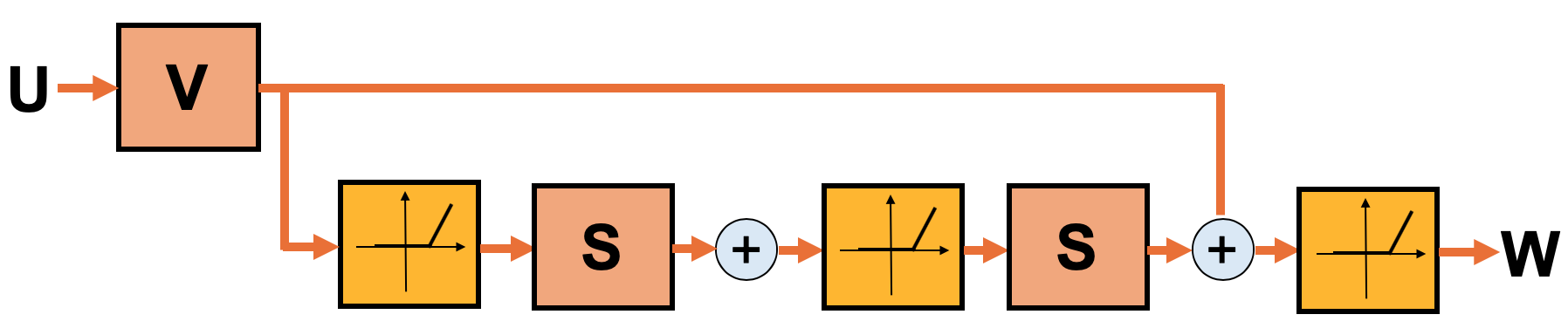}
& \hspace{0pt} 
\end{tabular}}
\caption{The block diagram shows learned ISTA by unfolding the iteration of ISTA for sparse NMF (2 times in this figure).}\label{fig:flowc} 
\end{figure}

\begin{equation}\label{eq:two_reg_NMF_single} 
\mathcal{E}_2 =  \frac{1}{2}\left\| {\mathbf{U} - \mathbf{VW}} \right\|_F^2 + \lambda\left\|\mathbf{W} \right\|_1 + \beta \mathrm{Tr}(\mathbf{WLW}^T),
\end{equation}
where $\lambda$ and $\beta$ denote the balancing parameters of the sparsity and graph regularizations, respectively, and $\mathbf{L} \in \mathbb{R}^{n\times n}$ represents the graph Laplacian matrix. The graph Laplacian is defined as $\mathbf{L=D-Q}$, where $\mathbf{Q}$ is a heat kernel weighting function associated with the input matrix $\mathbf{U}$, and the degree matrix $\mathbf{D}$ is a diagonal matrix whose entries are $\mathbf{D}_{jj}  = \sum\limits_l {\mathbf{Q}_{jl}}$. Minimizing the graph regularization term, $\mathrm{Tr}(\mathbf{WLW}^T)$, serves as a smoothing operator. In this work, the building block, $\mathbf{V}$, and the initial weighting map, $\mathbf{W}^{(0)}$, are first initialized using the work by~\cite{cai2010graph}. The ISTA method is then used to solve Eq.~(\ref{eq:two_reg_NMF_single}) for $\mathbf{W}$ as in Fig.~\ref{fig:flowc} and Algorithm~\ref{alg:thealg}:

\begin{algorithm}\label{alg:thealg}
 \textbf{Input:} motion feature matrix $\mathbf{U}$, building block $\mathbf{V}$ and initial weighting map $\mathbf{W}^{(0)}$ \\
 \For{h = 1 to H}{
 	$\mathbf{S}^{(h)} = \mathbf{W}^{(h-1)} + \frac{1}{c}\mathbf{V}^T (\mathbf{U}-\mathbf{V}\mathbf{W}^{(h-1)}) -\frac{\beta}{c}\mathbf{W}^{(h-1)}\mathbf{L}$ \\
    $\mathbf{W}^{(h)} = \mathbf{Soft}_{\lambda/c}(\mathbf{S}^{(h)})$
    } 
 \textbf{Return} $\mathbf{W}^{(H)}$ 
 \caption{ISTA to solve Eq.~(\ref{eq:two_reg_NMF_single})}
\end{algorithm}

Here 1/$c$, $h$, and $\mathbf{Soft}_{\alpha}(\mathbf{z})$ denote the step size, the ISTA iteration index, and the soft thresholding function with a threshold value $\lambda$/c, respectively. $\mathbf{Soft}_{\lambda/c}(\mathbf{z})$ is given by

\begin{equation}\label{eq:soft}
\mathbf{Soft}_{\lambda/c}(z_n) = \mathrm{sign}(z_{n}) \mathrm{max}(|z_{n}|-\lambda/c, 0).
\end{equation}

Because of the non-negative constraint imposed on $\mathbf{W}$, the soft-thresholding operation can be seen as a rectified linear unit (ReLU) activation function. It is worth noting that this minimization is equivalent to a fully connected layer, followed by ReLU activation, which bears structural similarity with the current deep neural network models. 

\subsubsection{Spectral Clustering}\label{sc} 

Once we obtain the weighting map from the ISTA method, we carry out clustering on the weighting map to partition the weighting map into disjoint subsets with high intra-cluster similarity, while maintaining low inter-cluster similarity via the eigen-structure of a data affinity graph. First, we construct an affinity matrix from the weighting map, which can be given by
\begin{equation}
\mathbf{A}(i,j) = \exp \left( { - \frac{{\left\| {w(i) - w(j)} \right\|_2 }}
{\sigma }} \right),
\end{equation}
where $w(i)$ denotes the $i$-th column vector of the weighting map $\mathbf{W}$, and $\sigma$ represents the scale factor. Then, spectral clustering~\citep{shi2000normalized} is carried out on the affinity matrix, followed by color-coding of each voxel within the tongue for visualization. 

\subsection{Deep Joint GS-NMF to Co-identify the Common and Subject-specific Functional Units}

\subsubsection{Construction of an average intensity and motion field atlas for a reference state from cine and tagged MRI}

An average intensity and four-dimensional (4-D) motion field atlas~\citep{woo2019speech} is built for a reference time frame from cine and tagged MRI. Due to large variability in speech movements across subjects, even for the same speech task, putting all the data into an atlas space is crucial to facilitate the comparison of subjects by standardizing varying tongue shape and size, and motion field for each subject. Toward this goal, a symmetric diffeomorphic registration using a cross-correlation (CC) similarity metric is used to construct the average intensity atlas~\citep{avants2011reproducible}. Let $\phi_i: \mathrm{\Omega}_{A} \to \mathrm{\Omega}_{i}$ denote the diffeomorphic transformation between the volume of the $i$-th subject and the atlas volume. Then, all the motion tracking results are mapped to the atlas space using the following transformation:

\begin{equation}\label{mapping}
\tilde\varphi_{i,j} = \phi_i \circ {\varphi_{i,j}} \circ \phi_i^{-1},
\end{equation}
where $\tilde\varphi_{i,j}$ represents a motion field from the reference time frame to the $j$-th time frame of the $i$-th subject transformed in the atlas space. This Lagrangian configuration allows us to anchor the root of the motion fields in the material coordinates, thereby allowing for all motion features to be mapped back to the static anatomy~\citep{woo2019speech}. Further, in order to achieve accurate time alignment across subjects, we visually identify the critical time instants (/\textschwa/, /s/, /u/, and /k/) of all the subjects from their imaging data and align those instants at the same time positions. The data between those critical time positions are then interpolated based on the original imaging data and spread over an evenly distributed time grid.

\subsubsection{Deep Joint GS-NMF}

Once we establish the atlas and put all the data in the atlas space, we form a feature matrix to identify functional units~\citep{woo2018sparse}. Let $\mathbf{U}_{i}$ denote an input non-negative feature matrix of the $i$-th subject, consisting of the magnitude and angle derived from displacements as in Eq.~(\ref{eq:feature}). The objective function of GS-NMF for identifying individual functional units is defined as 

\begin{equation}\label{eq:two_reg_NMF} 
\mathcal{E}_3 =  \frac{1}{2}\left\| {\mathbf{U}_{i} - \mathbf{V}_{i}\mathbf{W}_{i}} \right\|_F^2 + \lambda\left\|\mathbf{W}_{i} \right\|_1 + \beta \mathrm{Tr}(\mathbf{W}_{i}\mathbf{L}_{i}\mathbf{W}_{i}^T),
\end{equation}
where $\mathbf{V}_{i}$, $\mathbf{W}_{i}$, and $\mathbf{L}_{i}$ represent building blocks, their weighting map, and the graph Laplacian matrix for the $i$-th subject, respectively, and $\lambda$ and $\beta$ are weighting parameters for sparsity and graph regularizations, respectively. In order to identify the common weighting map, the following loss function, a measure of disparity between each weighting map and the common weighting map, is defined as
\begin{equation}
\mathcal{E}_4 = \frac{1}{2}\left\| {{\mathbf{W}_{i}} - {\mathbf{W}^{*}}} \right\|_F^2,
\end{equation}
where $\mathbf{W}^*$ represents the common weighting map.

The overall objective function to find the building blocks, subject-specific weighting maps, and common weighting map is then defined as

\begin{equation}\label{eq:final_NMF} 
\begin{array}{l}
\mathcal{O}_2 = \sum\limits_{i=1}^{N} {\mathcal{E}_3}  + \sum\limits_{i = 1}^{N} {{\gamma}\mathcal{E}_4} \\
\mathrm{s.t.} \forall i \in \{1,\cdot\cdot\cdot, {N}\},{\mathbf{V}_{i}},{\mathbf{W}_{i}},{\mathbf{W}^*} \ge 0
\end{array}
\end{equation}
where $N$ denotes the number of subjects and $\gamma$ represents a weighting parameter between the GS-NMF reconstruction error and the disparity term incorporating the common weighting map. 

The objective function is optimized via an iterative and alternative ISTA update scheme as in Algorithm~\ref{alg:algorithm_2}: 

\begin{algorithm}\label{alg:algorithm_2} 
 \textbf{Input:} motion feature matrix $\mathbf{U}_{i}$, building block $\mathbf{V}_{i}$, initial weighting map $\mathbf{W}_{i}^{(0)}$ initialized by~\cite{cai2010graph}, and initial common weighting map initialized by Eq.~(\ref{eq:common}) \\
 \textbf{Output:} $\mathbf{W}_{i}^{(H)}$, $\mathbf{W}^{*}$ \\
 \For{i = 1 to $N$}{

 \For{h = 1 to H}{
 	Solving for $\mathbf{S}_{i}^{(h)}$ using Eq.~(\ref{eq:eq_s1}) by fixing $\mathbf{V}_{i}$ and $\mathbf{W}^{*(h)}$ \\
 	Solving for $\mathbf{W}_{i}^{(h)}$ using Eq. (\ref{eq:eq_s2}) by fixing $\mathbf{V}_{i}$ and $\mathbf{W}^{*(h)}$ \\
 } 
 }
  Solving for $\mathbf{W}^{*}$ using Eq. (\ref{eq:common}) by fixing $\mathbf{W}_{i}^{(h)}$

 \textbf{Return} $\mathbf{W}_{i}^{(H)}$, $\mathbf{W}^{*}$
 \caption{ISTA to solve Eq.~(\ref{eq:final_NMF})}
\end{algorithm}

\begin{equation}\label{eq:eq_s1}
\begin{split}
\mathbf{S}_{i}^{(h)} = \mathbf{W}_{i}^{(h-1)} + \frac{1}{c}[\mathbf{V}_{i}^T (\mathbf{U}_{i}-\mathbf{V}_{i}\mathbf{W}_{i}^{(h-1)}) \\
-\gamma({{\mathbf{W}_{i}^{(h-1)}} - {\mathbf{W}^{*}}})]-\frac{\beta}{c}\mathbf{W}_{i}^{(h-1)}\mathbf{L}_{i}
\end{split}
\end{equation}

\begin{equation}\label{eq:eq_s2}
\mathbf{W}_{i}^{(h)} = \mathbf{Soft}_{\lambda/c}(\mathbf{S}_{i}^{(h)})
\end{equation}

\begin{equation}\label{eq:common}
{\mathbf{W}^{*}} = \frac{{\sum\nolimits_{i = 1}^N {{\alpha_i}{\mathbf{W}_{i}^{(h)}}} }}{{\sum\nolimits_{i = 1}^N {{\alpha_i}} }},
\end{equation}
where 1/$c$, $h$, and $\mathbf{Soft}_{\lambda/c}(\mathbf{z})$ denote the step size, the ISTA iteration index, and the soft thresholding function with a threshold value $\lambda$/c as in Eq.~(\ref{eq:soft}), respectively. 

\subsubsection{Spectral Clustering}

The final subject-specific and common functional units are then obtained by applying spectral clustering to the weighting map for each subject and the common weighting map as described in Sec.~\ref{sc}.

\subsection{Complexity Analysis}
In this subsection, we discuss the time complexity of our approach in comparison to the prior works.~For simplicity, we assume that the time complexity of multiplication of two matrices---e.g., a $p$$\times$$q$ matrix and a $q$$\times$$r$ matrix---is $O$($pqr$)~\citep{cormen2009introduction}.~The time complexity required for evaluating the recursive formula of our approach is $O$($N$($Hmnk$ + $n^2m$)), where $N$ is the number of subjects and $H$ is the ISTA iteration number. The time complexities for GS-NMF-S and ISTA-S-NMF-S per subject are $O$($t_{p}$$mnk$ + $n^2m$)) and $O$($Hmnk$), respectively, where $t_{p}$ is the iteration number for the multiplicative updates. Thus, our approach has a time complexity similar to the prior works, while achieving superior clustering performance.

\section{Experimental Results}\label{sec:result}

In this section, we validate our approach described above on 2D plus time and 3D plus time synthetic motion data first because of the lack of ground truth in the \textit{in vivo} tongue motion data. We then use \textit{in vivo} tongue motion data obtained by tagged MRI to determine the common and subject-specific functional units. The clustering results are typically evaluated by comparing the label of each data point computed by methods against the ground truth label. Both the accuracy (AC) and the normalized mutual information (NMI) have been widely used to evaluate the clustering performance~\citep{cai2010graph,ghasedi2017deep}. Specifically, given points defined within the tongue $x_i$, let $s_i$ and $g_i$ be the label identified by each method and the ground truth label, respectively. Then, AC is defined as:

\begin{equation}\label{eq:ac}
\mathrm{AC} = \frac{{\sum\nolimits_{i = 1}^n {\delta(g_i, map(s_i))}}} {{n}} \times 100,
\end{equation}
where $n$ denotes the total number of points, $map (s_i)$ represents the mapping function that aligns each cluster label $s_i$ with the ground truth label $g_i$ via the Kuhn-Munkres algorithm~\citep{lovasz2009matching}, and $\delta(a, b)$ is given by 

\begin{equation}
\delta(a, b) = \begin{cases}
1 & \text{if} \; a=b\\
0 &\text{otherwise}.
\end{cases}
\end{equation} 

Let $A$ and $B$ denote the set of clusters computed from each method and ground truth, respectively, and the mutual information metric is given by

\begin{equation}\label{eq:eq_mi}
\mathrm{MI} (A, B) = \sum_{a_i \in A , b_i \in B} p(a_i, b_i) \cdot log_2\frac{p(a_i, b_i)}{p(a_i)\cdot p(b_i)},
\end{equation}
where $p(a_i)$ and $p(b_i)$ represent the probabilities that a point from the whole points belongs to the clusters $a_i$ and $b_i$, respectively, and $p(a_i, b_i)$ represents the joint probability that the selected point simultaneously belongs to the clusters $a_i$ and $b_i$. Then, NMI is defined as follows:

\begin{equation}\label{eq:eq_nmi}
\mathrm{NMI} (A, B) = \frac{ \mathrm{MI}(A, B)}{max(H(A), H(B))} \times 100,
\end{equation}\label{eq:eq_nmi}~where $H(A)$ and $H(B)$ denote the entropies of $A$ and $B$, respectively. Note that the range of NMI value is from 0 to 100, where 100 means the two set of clusters are the same.

%% Table 1 %%%%%%%%%%%%%%%%%%%%%%%%%%%%%%%%%%%%%%%%%%%%%%%%%%%%%%%
\begin{table*}[h]
\caption{Comparison methods and their characteristics}
\centering
%\scriptsize{ 
\begin{tabular}{c|c} \hline
 \hline 
  Method & Key characteristics \\
   \hline \hline
  G-NMF-S~\citep{cai2010graph} & Shallow NMF, sparse regularization \\
  GS-NMF-S~\citep{woo2018sparse} & Shallow NMF, both sparse and graph regularizations \\
  ISTA-S-NMF-S (a variant of~\cite{gregor2010learning}) & Deep NMF, sparse regularization \\
  ISTA-GS-NMF-S (proposed) & Deep NMF, both sparse and graph regularizations \\
 \hline
\end{tabular}\label{table:name}
\end{table*}
%%%%%%%%%%%%%%%%%
%%%%%%%%%%%%%%%%%%%%%%%%%%%%%%%%%%%%%%%%%%%%%%%%%%%%%%%%
\def\FigureHeight{65mm}
\begin{figure}[t]
 \center{
 \begin{tabular}{c@{ }c}
   \includegraphics[trim=0mm 0mm 0mm
0mm,clip=true,height=\FigureHeight]{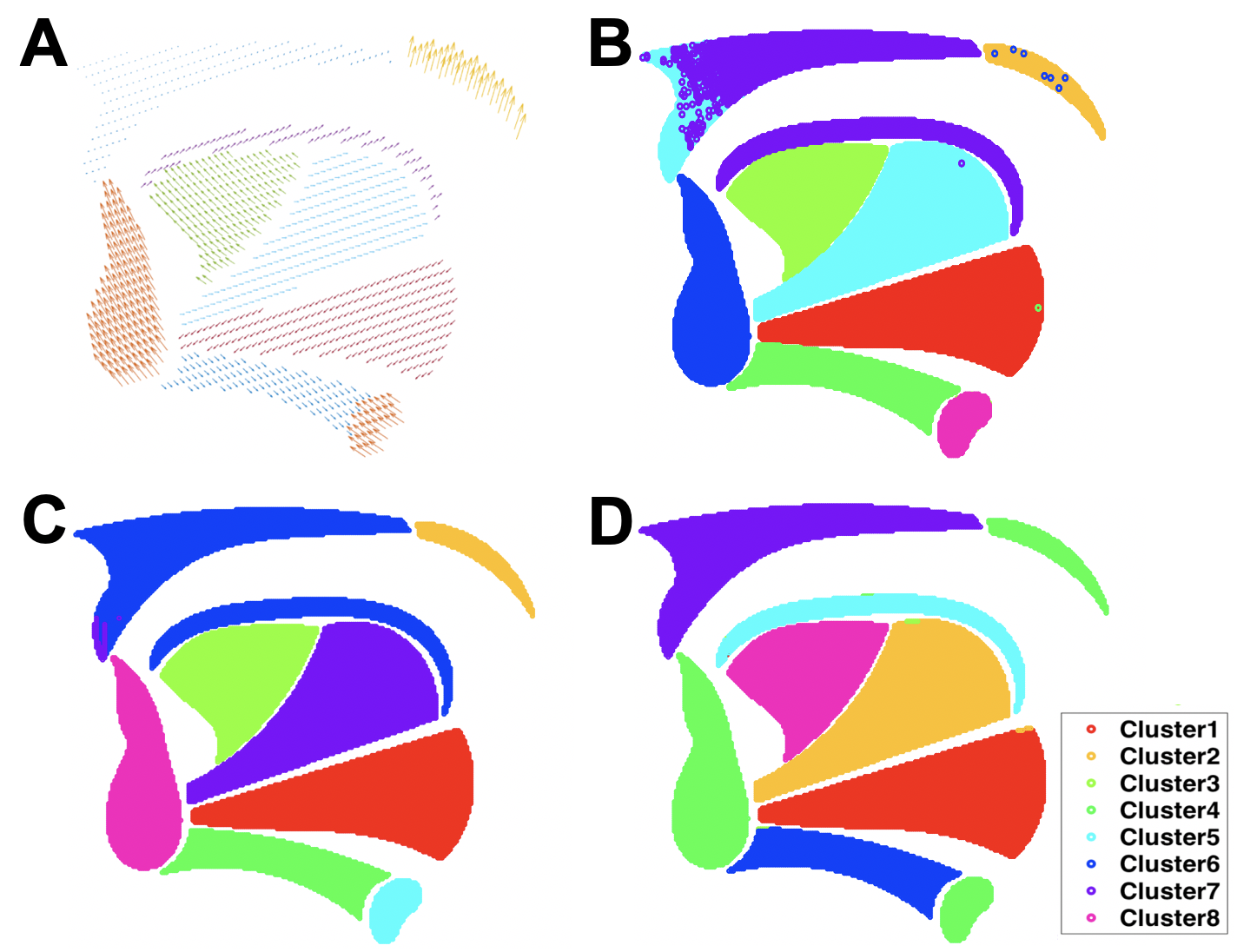}
& \hspace{0pt} 
\end{tabular}}
\caption{Illustration of 2D synthetic tongue motion simulation: (A) 2D displacement field, (B) the result using GS-NMF-S, (C) the result using ISTA-S-NMF-S, and (D) the result using our proposed approach. The different color represents different class labels.}\label{fig:flowc2d} 
\end{figure}

%%%%%%%%%%%%%%%%%%%%%%%%%%%%%%%%%%%%%%%%%%%%%%%%%%%%%%%%
\def\FigureHeight{57mm}
\begin{figure*}[t]
 \center{
 \begin{tabular}{c@{ }c}
   \includegraphics[trim=0mm 0mm 0mm
0mm,clip=true,height=\FigureHeight]{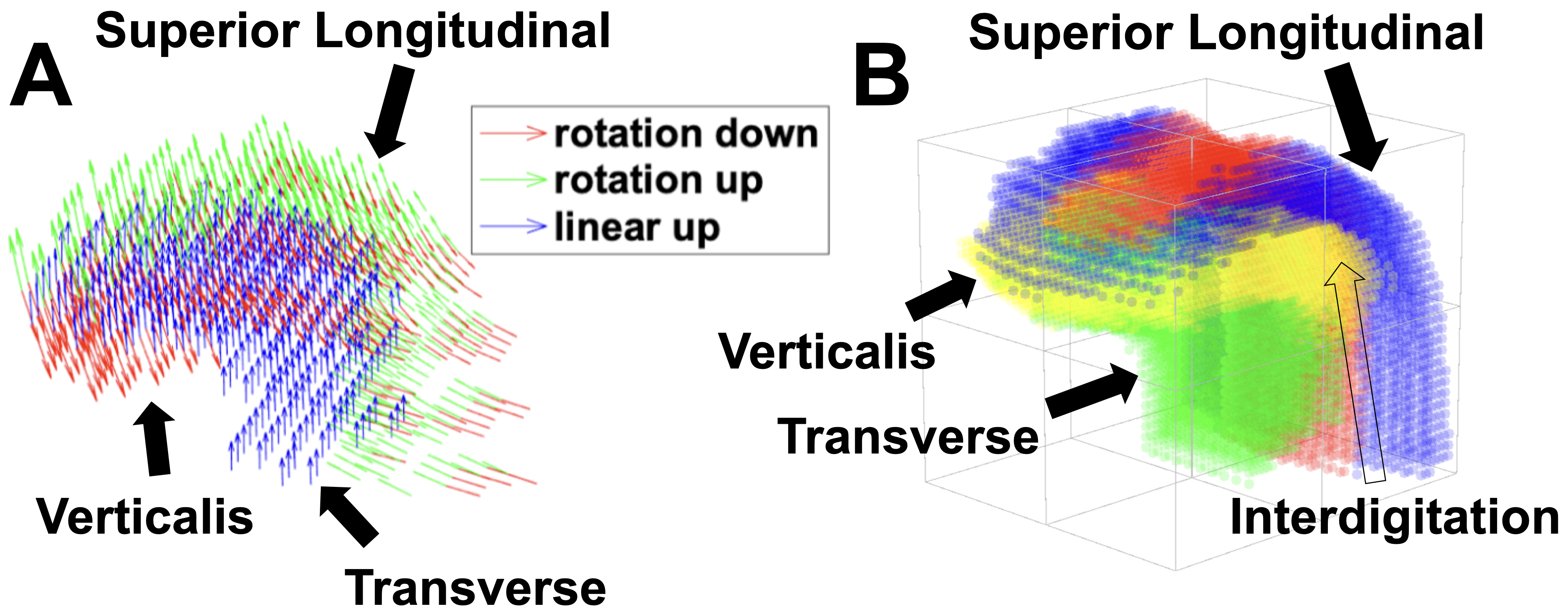}
& \hspace{0pt} 
\end{tabular}}
\caption{Illustration of 3D synthetic tongue motion simulation (data 1): (A) 3D displacement field and (B) the ground truth labels.}\label{fig:flowc3d_1} 
\end{figure*}

\subsection{Experiments Using Synthetic Tongue Motion Data}

Our strategy for quantitative evaluation was to use the proposed method and the comparison methods (see Table~\ref{table:name}), including graph-regularized NMF + spectral clustering (G-NMF-S), graph-regularized sparse NMF + spectral clustering (GS-NMF-S)~\citep{woo2018sparse}, and ISTA for sparse NMF + spectral clustering (ISTA-S-NMF-S) to extract groupings from a synthetic displacement field composed of known areas representative of functional units. We then analyzed the difference between the grouping as outputted by the methods against the known distribution. We constructed simulated 2D and 3D displacement fields based on a tongue geometry derived from a vocal tract atlas previously developed~\citep{woo2015high, stone2018structure}. The 2D displacement fields were based on the areas illustrated in Fig.~\ref{fig:flowc2d} and included Lagrangian displacements of heterogeneous magnitude representative of vertical, horizontal movement, and rotations in one deformed configuration. Table~\ref{table:2d} lists numerical comparisons between G-NMF-S, GS-NMF-S~\citep{woo2018sparse}, ISTA-S-NMF-S, and the proposed approach (ISTA-GS-NMF-S). The results indicated that our approach surpassed the comparison methods in our 2D experiments. In our experiments, we chose $\lambda$=500, $c$=100, $\beta$=0, $H$=10, and $\sigma$=0.07 for ISTA-S-NMF-S and $\lambda$=500, $c$=100, $\beta$=0.05, $H$=10, and $\sigma$=0.07 for our approach. These parameters were chosen to empirically maximize the clustering performance.

The 3D displacement fields included two temporal sequences of Lagrangian motion across 11 time frames each. The first dataset included spatially heterogeneous displacement fields as displayed in Fig.~\ref{fig:flowc3d_1}. The displacements were distributed based on the location of the verticalis (V), superior longitudinal (SL), and Transverse (T), which were defined using the vocal tract atlas for each time frame. We note that in the first dataset, the V and SL as well as the V and T muscles interdigitated with each other, respectively. Thus, we had a total of four ground truth labels in our quantitative evaluation. In addition, the V and SL muscles were rotated downward and upward, respectively, while the T muscle was translated upward in the course of 11 time frames (see Fig.~\ref{fig:flowc3d_1}). The second dataset also had composite Lagrangian displacement fields from 11 time frames as displayed in Fig.~\ref{fig:flowc3d_2}. We used the composite displacement field of genioglossus (GG), T, and geniohyoid (GH), which also were defined using the vocal tract atlas. We note that in the second dataset, the GG and T interdigitated with each other, and therefore we had a total of four ground truth labels in our quantitative evaluation. The GG and T muscles were rotated downward and upward, respectively, while the GH muscle was translated upward in the course of 11 time frames (see Fig.~\ref{fig:flowc3d_2}). The clustering outcomes using different methods are listed in Table~\ref{table:3d}, demonstrating that our approach achieved an accuracy level comparable or better than the comparison methods. In our experiments, for the first dataset, we chose $\lambda$=890, $c$=55, $\beta$=0.03, $H$=49, and $\sigma$=0.05 for ISTA-S-NMF-S and $\lambda$=890, $c$=55, $\beta$=0.03, $H$=49, and $\sigma$=0.05 for our approach. For the second dataset, we chose $\lambda$=800, $c$=100, $\beta$=0, $H$=50, and $\sigma$=0.03 for ISTA-S-NMF-S and $\lambda$=800, $c$=100, $\beta$=0.05, $H$=50, and $\sigma$=0.03 for our approach. These parameters were chosen to empirically maximize the clustering performance as shown in Figs.~\ref{fig:param1} and~\ref{fig:param2}. In Fig.~\ref{fig:param1}, in the case of the first dataset, for $H$, there was a local maxima, but in the case of the second dataset, in Fig.~\ref{fig:param2}, after 20 iterations, our proposed approach converged to a global maxima, the perfect score, which appears to be a special case. The effects of $\lambda$ and $\beta$ on the clustering performance of 3D tongue simulation data are shown in Table~\ref{table:3d_param}. Our simulation study using 2D data shows that our approach using both regularizations achieved better performance, whereas our simulation study using 3D data shows that our approach performed on par with ISTA-S-NMF-S.

%% Table 1 %%%%%%%%%%%%%%%%%%%%%%%%%%%%%%%%%%%%%%%%%%%%%%%%%%%%%%%
\begin{table*}[h]
\caption{Clustering Performance of 2D Tongue Simulation Data}
\centering
%\scriptsize{ 
\begin{tabular}{c||c|c|c|c} \hline
 \hline 
  2D Tongue ($\%$) & G-NMF-S & GS-NMF-S & ISTA-S-NMF-S & ISTA-GS-NMF-S \\
   \hline \hline
  AC & 85.74 & 86.15 & 91.23 & \textbf{97.53} \\
 \hline 
  NMI & 89.30 & 89.16 & 94.30 & \textbf{97.84} \\
 \hline
\end{tabular}\label{table:2d}
\end{table*}
%%%%%%%%%%%%%%%%%

%%%%%%%%%%%%%%%%%%%%%%%%%%%%%%%%%%%%%%%%%%%%%%%%%%%%%%%%
\def\FigureHeight{54mm}
\begin{figure*}[t]
 \center{
 \begin{tabular}{c@{ }c}
   \includegraphics[trim=0mm 0mm 0mm
0mm,clip=true,height=\FigureHeight]{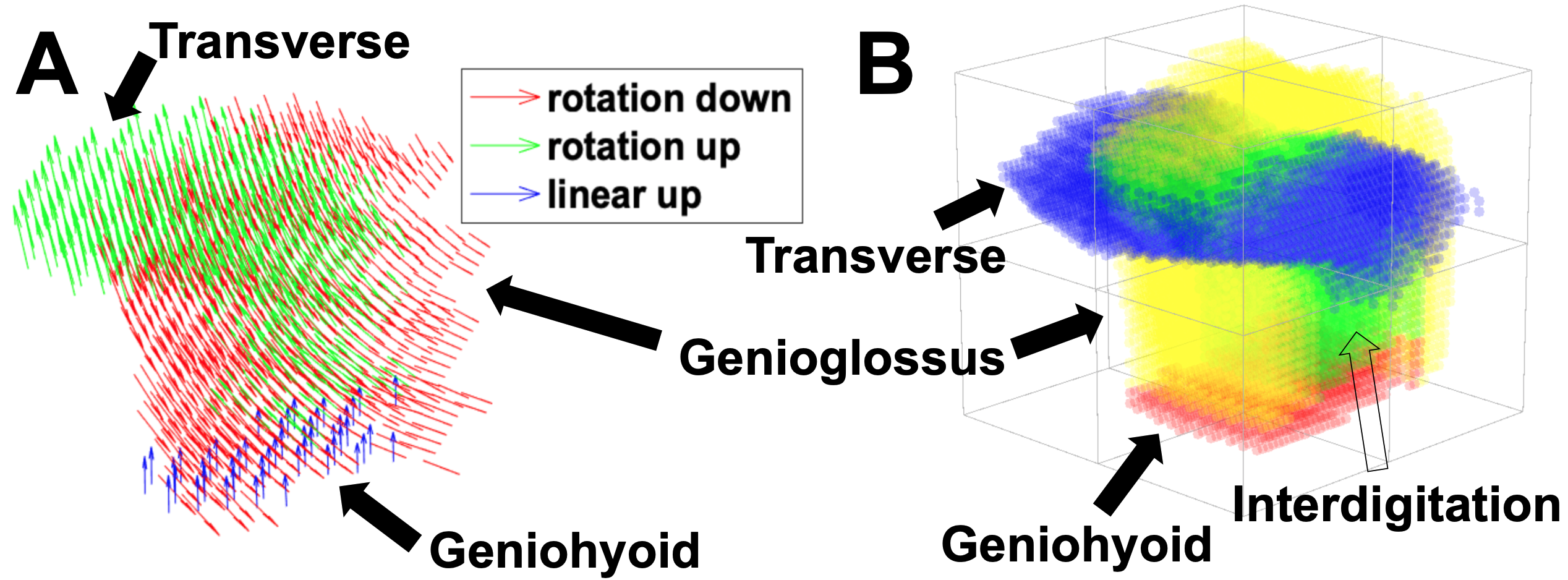}
& \hspace{0pt} 
\end{tabular}}
\caption{Illustration of 3D synthetic tongue motion simulation (data 2): (A) 3D displacement field and (B) the ground truth labels.}\label{fig:flowc3d_2} 
\end{figure*}

%%%%%%%%%%%%%%%%%%%%%%%%%%%%%%%%%%%%%%%%%%%%%%%%%%%%%%%%
\def\FigureHeight{64mm}
\begin{figure}[t]
 \center{
 \begin{tabular}{c@{ }c}
   \includegraphics[trim=0mm 0mm 0mm
0mm,clip=true,height=\FigureHeight]{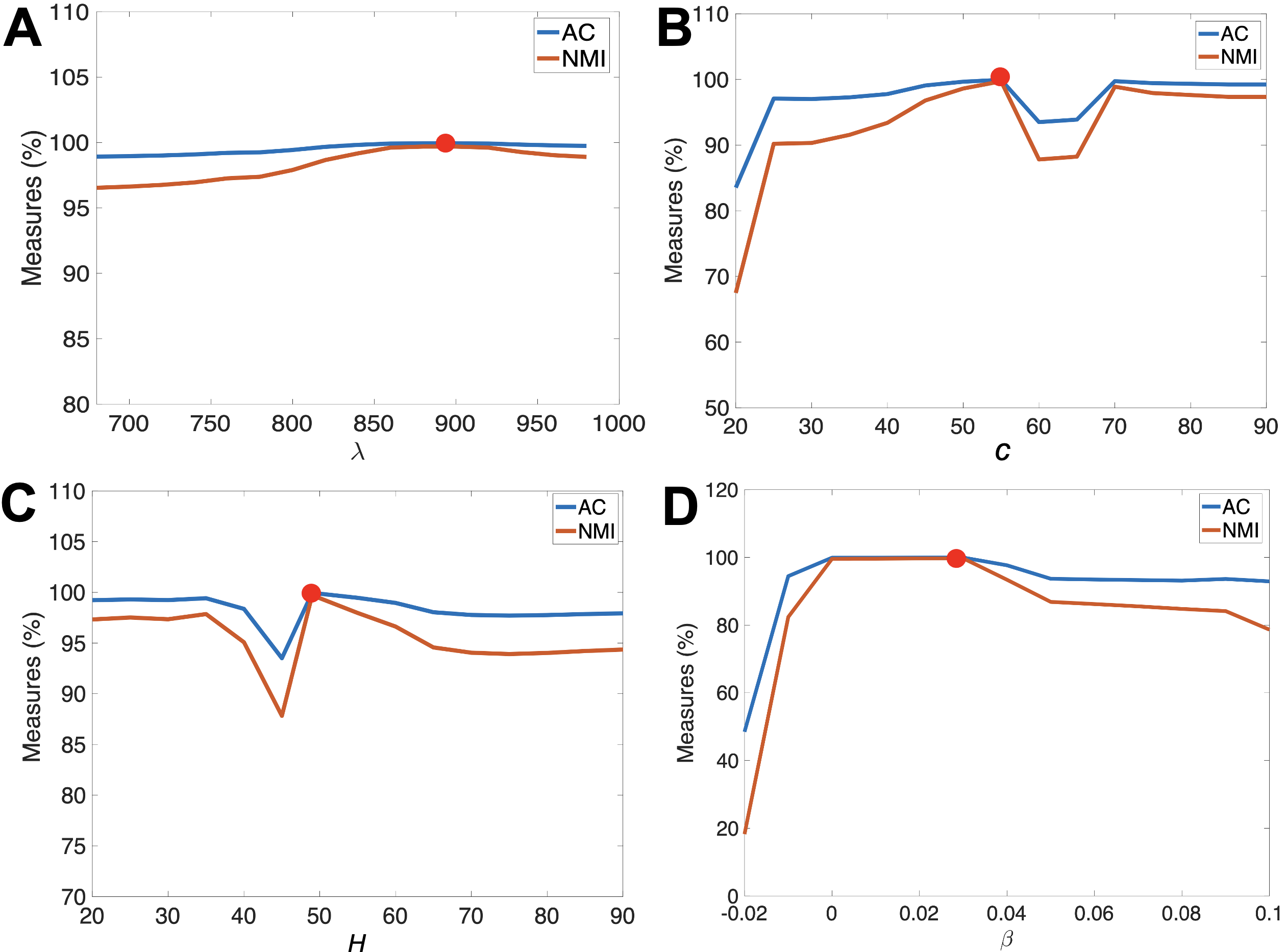}
& \hspace{0pt} 
\end{tabular}}
\caption{The performance of the proposed approach with respect to the parameters for the first dataset: (A) $\lambda$, (B) $c$, (C) $H$, and (D) $\beta$. The red dot indicates the optimal parameter used for our simulation.}\label{fig:param1} 
\end{figure}
%%%%%%%%%%%%%%%%%%%%%%%%%%%%%%%%%%%%%%%%%%%%%%%%%%%%%%%%
\def\FigureHeight{64mm}
\begin{figure}[t]
 \center{
 \begin{tabular}{c@{ }c}
   \includegraphics[trim=0mm 0mm 0mm
0mm,clip=true,height=\FigureHeight]{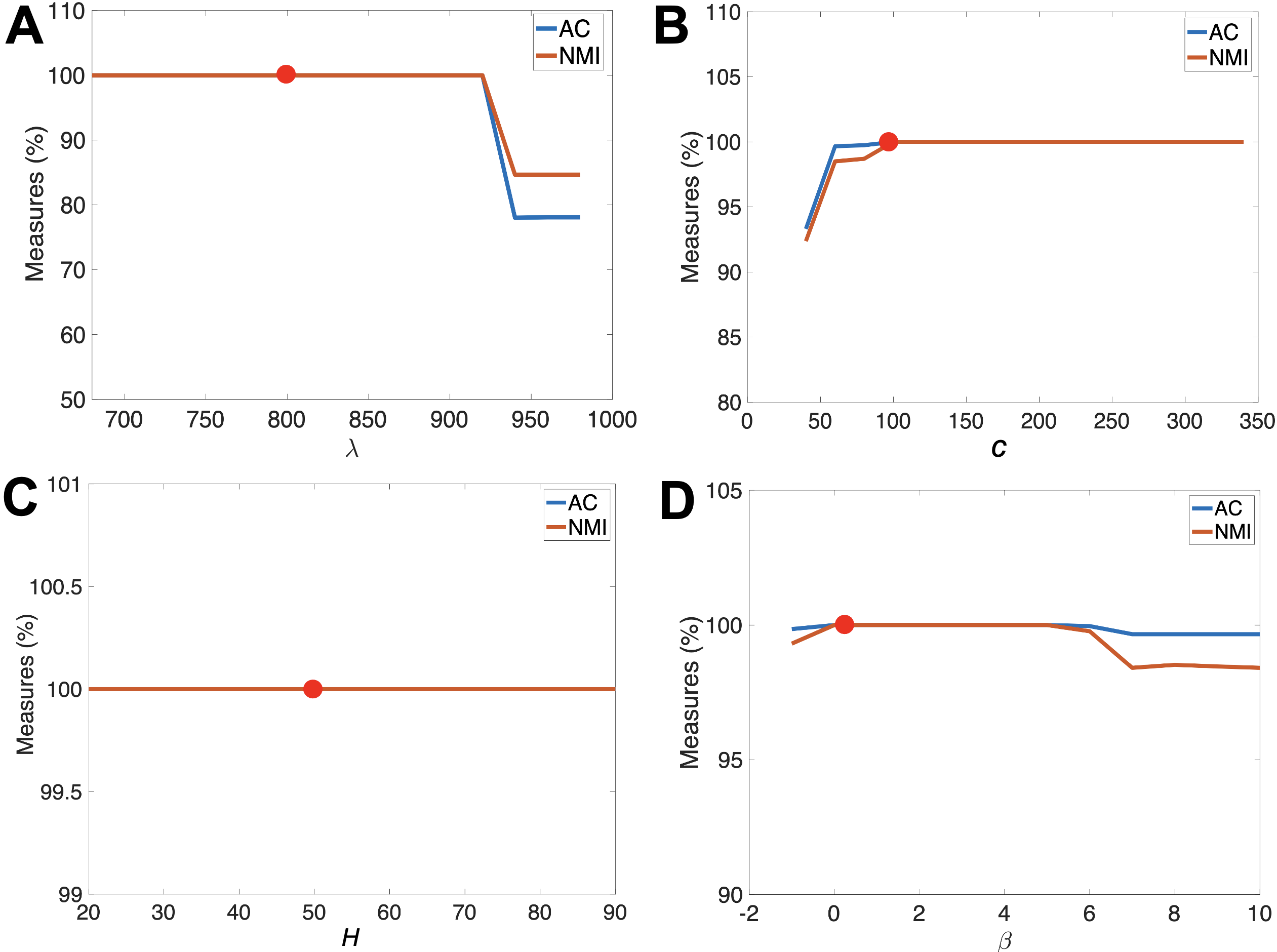}
& \hspace{0pt} 
\end{tabular}}
\caption{The performance of the proposed approach with respect to the parameters for the second dataset: (A) $\lambda$, (B) $c$, (C) $H$, and (D) $\beta$. The red dot indicates the optimal parameter used for our simulation.}\label{fig:param2} 
\end{figure}

%% Table 1 %%%%%%%%%%%%%%%%%%%%%%%%%%%%%%%%%%%%%%%%%%%%%%%%%%%%%%%
\begin{table*}[h]
\caption{Clustering Performance of 3D Tongue Motion Simulation Data}
\centering
\begin{tabular}{c||c|c|c|c} \hline
 \hline 
  Data 1 ($\%$) & G-NMF-S & GS-NMF-S & ISTA-S-NMF-S & ISTA-GS-NMF-S \\
   \hline \hline
  AC & 98.57 & 98.58 & 99.92 & \textbf{99.95} \\
 \hline 
  NMI & 95.70 & 95.73 & 99.58 & \textbf{99.72} \\
 \hline  \hline
 Data 2 ($\%$) & G-NMF-S & GS-NMF-S & ISTA-S-NMF-S & ISTA-GS-NMF-S \\
   \hline \hline
   AC & 99.37 & 99.36 & \textbf{100} & \textbf{100} \\
 \hline 
  NMI & 98.00 & 97.99 & \textbf{100} & \textbf{100} \\
 \hline
\end{tabular}\label{table:3d}
\end{table*}
%%%%%%%%%%%%%%%%%

%% Table 1 %%%%%%%%%%%%%%%%%%%%%%%%%%%%%%%%%%%%%%%%%%%%%%%%%%%%%%%
\begin{table*}[h]
\caption{Impact of Each Parameter on the Clustering Performance of 3D Tongue Motion Simulation Data}
\centering
\begin{tabular}{c||c|c|c} \hline
 \hline 
  Data 1 ($\%$) & $\lambda$=0  & $\beta$=0 & $\lambda$=0 and $\beta$=0  \\
   \hline \hline
  AC & 36.85 & 99.92 & 81  \\
 \hline 
  NMI & 0.62 & 99.58 & 74.1  \\
 \hline  \hline
 Data 2 ($\%$) & $\lambda$=0  & $\beta$=0 & $\lambda$=0 and $\beta$=0  \\
   \hline \hline
   AC & 37.36 & 100 & 99.38  \\
 \hline 
  NMI & 1.78 & 100 & 98.05  \\
 \hline
\end{tabular}\label{table:3d_param}
\end{table*}
%%%%%%%%%%%%%%%%%

%%%%%%%%%%%%%%%%%%%%%%%%%%%%%%%%%%%%%%%%%%%%%%%%%%%%%%%%
\def\FigureHeight{106mm}
\begin{figure*}
 \center{
 \begin{tabular}{c@{ }c}
   \includegraphics[trim=0mm 0mm 0mm
0mm,clip=true,height=\FigureHeight]{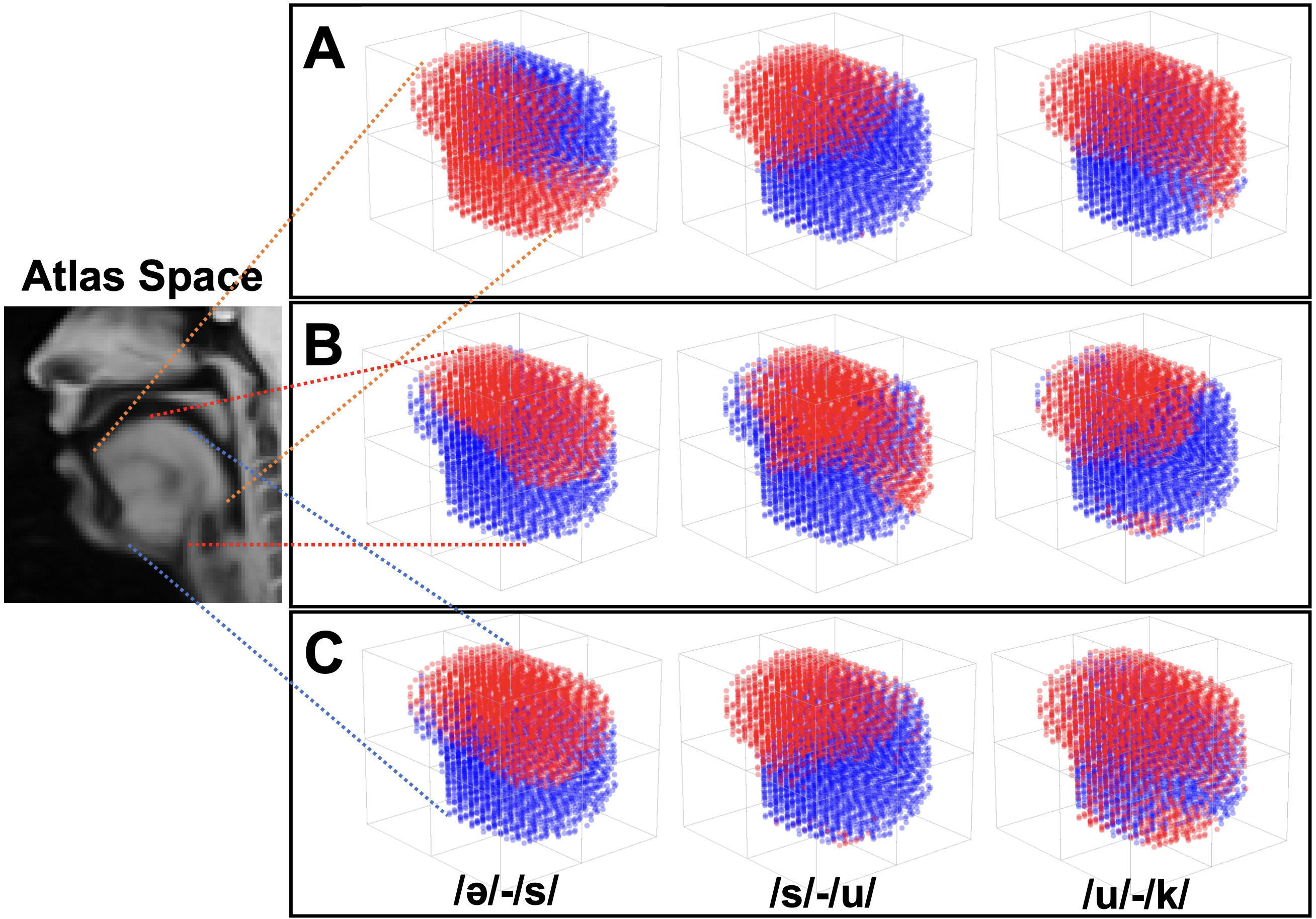}
& \hspace{0pt} 
\end{tabular}}
\caption{Illustration of (A) the common functional units (2 units) identified using our proposed approach, (B) two functional units identified using the prior work~\citep{woo2018sparse}, and (C) two functional units identified using our proposed method for the transitions of /\textschwa/-/s/, /s/-/u/, and /u/-/k/, respectively.}\label{fig:two_units}
\end{figure*}
%%%%%%%%%%%%%%%%%%%%%%%%%%%%%%%%%%%%%%%%%%%%%%%%%%%%%%%%

\def\FigureHeight{106mm}
\begin{figure*}
 \center{
 \begin{tabular}{c@{ }c}
   \includegraphics[trim=0mm 0mm 0mm
0mm,clip=true,height=\FigureHeight]{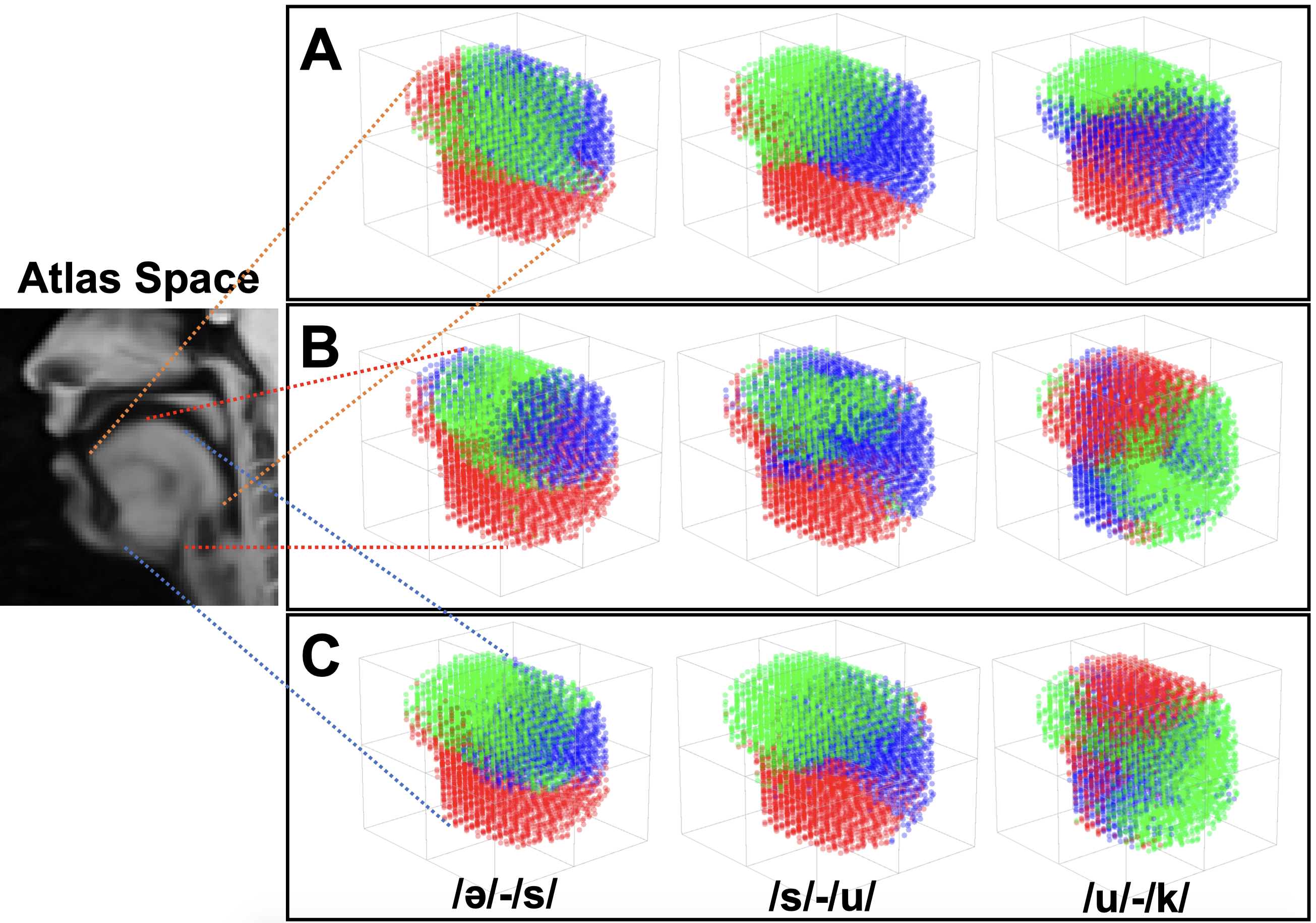}
& \hspace{0pt} 
\end{tabular}}
\caption{Illustration of (A) the common functional units (3 units) identified using our proposed approach, (B) three functional units identified using the prior work~\citep{woo2018sparse}, and (C) three functional units identified using our proposed method for the transitions of /\textschwa/-/s/, /s/-/u/, and /u/-/k/, respectively.}\label{fig:three_units}
\end{figure*}

\def\FigureHeight{74mm}
\begin{figure*}[t]
 \center{
 \begin{tabular}{c@{ }c}
   \includegraphics[trim=0mm 0mm 0mm
0mm,clip=true,height=\FigureHeight]{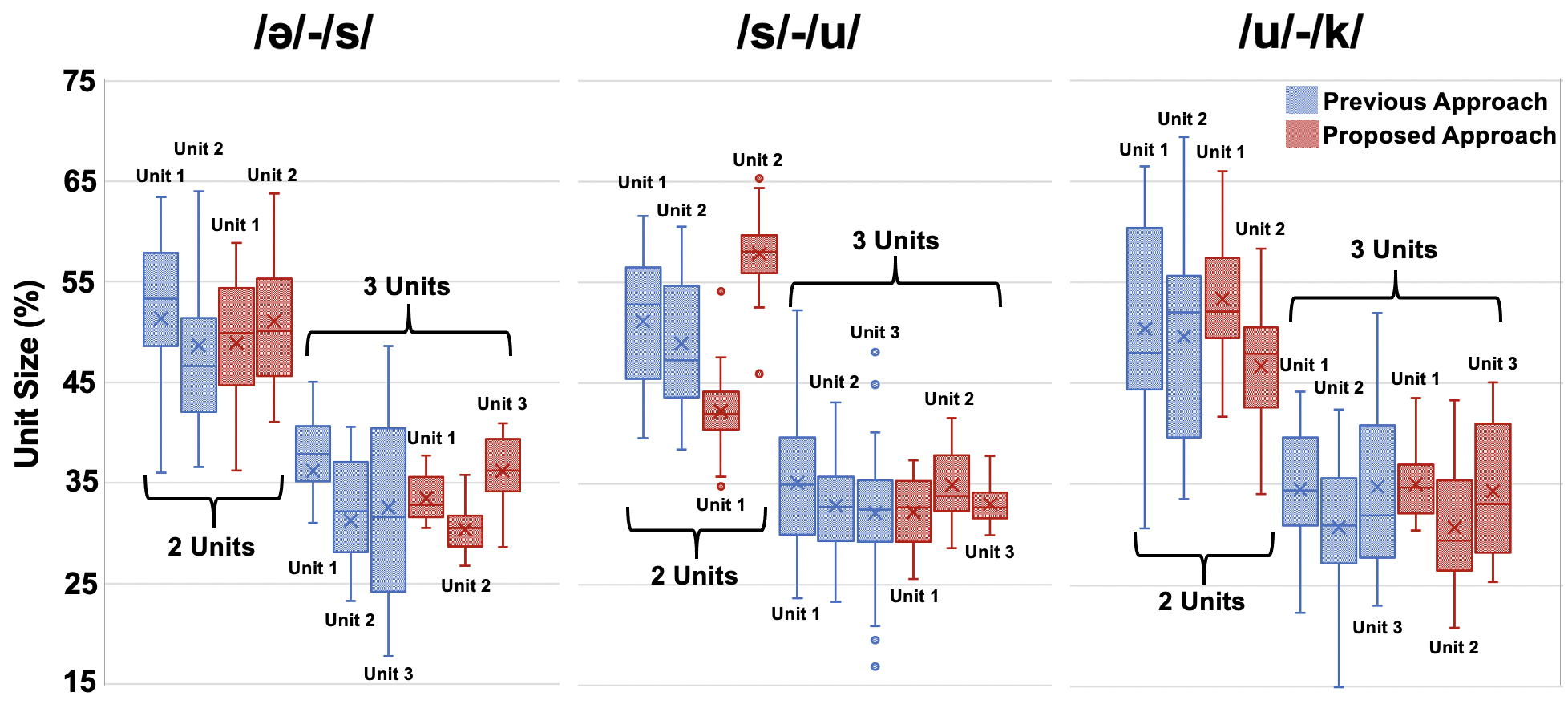}
& \hspace{0pt} 
\end{tabular}}
\caption{The comparison of the sizes of the three functional units using our approach and the previous approach~\citep{woo2018sparse} for the transitions of /\textschwa/-/s/, /s/-/u/, and /u/-/k/.}\label{fig:box} 
\end{figure*}

\subsection{Experiments Using \textit{In Vivo} Tongue Motion Data}

We applied our proposed framework to a cohort of healthy subjects with a simple word ``a souk'' to identify both the common and subject-specific functional units in the atlas space. We first transformed all the motion fields into the atlas space. Second, we extracted the motion quantities, including the magnitude and angle of the motion trajectories and constructed an input spatiotemporal matrix containing 18 healthy subjects. Finally, we scaled the matrix, which was then inputted into our deep joint sparse NMF framework described above. The F-test was used to compare the variability of the sizes of the identified functional units from different approaches with a level of significance set at $p$$<$0.05. In addition, to test the normality of the sizes of the identified functional units, we performed the Anderson-Darling test~\citep{scholz1987k}. In all the experiments below, we chose $\lambda$=800, $c$=600, $\gamma$=20, $H$=100, and $\beta$=0.03 that are consistent with our experiments using the 3D tongue simulator.

Figs.~\ref{fig:two_units} and~\ref{fig:three_units} show two and three unit cluster representations of the common functional units, subject-specific functional units (subject 7 in Table~\ref{table:subject}) identified using the prior work~\citep{woo2018sparse}, and subject-specific functional units identified using our proposed approach of three distinct phonemes, including (1) /\textschwa/-/s/, (2) /s/-/u/, and (3) /u/-/k/ from ``a souk.'' 

For the transition of /\textschwa/ to /s/, the two functional units in Fig.~\ref{fig:two_units}(A) showed that the tip and base of the tongue are clustered together, which represents forward/upward motion, while the posterior tongue was clustered as a separate unit, which represents forward motion. The results from the proposed approach in Figs.~\ref{fig:two_units}(B) and~\ref{fig:three_units}(B) showed clearer divisions between the units, thereby yielding more interpretable results in relation to the common functional units than the previous approach as visually assessed. The three functional units in Fig.~\ref{fig:three_units}(A) created clear divisions between the tongue base, tip, and posterior tongue. For the transition of /s/ to /u/, the two functional units (Fig.~\ref{fig:two_units}(A)) showed divisions between the anterior and posterior tongue. The three functional units (Fig.~\ref{fig:three_units}(A)) further formed clear divisions between the anterior, base, and posterior tongue. For the transition of /u/ to /k/, the upper tongue was clustered, since the tongue body was elevated, while the base and the body of the tongue were divided into separate units.

Notably, the functional units identified using our approach need to be interpreted in relation to the common functional units. More specifically, in Fig.~\ref{fig:two_units}(B) and Fig.~\ref{fig:three_units}(B), except for /\textschwa/-/s/, the functional units identified using the proposed approach look similar to the common functional units in Fig.~\ref{fig:two_units}(A) and Fig.~\ref{fig:three_units}(A). The functional units identiﬁed using the previous approach in Fig.~\ref{fig:two_units}(C) and Fig.~\ref{fig:three_units}(C), however, looks quite different. Fig.~\ref{fig:two_units}(C) and Fig.~\ref{fig:three_units}(C) appear to use a different strategy from the common functional units and the functional units in Fig.~\ref{fig:two_units}(B) and Fig.~\ref{fig:three_units}(B). For example, in the /\textschwa/-/s/ motion, Fig.~\ref{fig:two_units}(C) and Fig.~\ref{fig:three_units}(C) have a bilateral difference in the upper tongue; the midline (green) is different from the lateral (blue), reﬂecting the development of a midline groove as the tongue moves into /s/. In Fig.~\ref{fig:two_units}(B) and Fig.~\ref{fig:three_units}(B), the groove development may be more subtle, i.e., smaller, or already present in /\textschwa/. In the /s/-/u/ motion, Fig.~\ref{fig:two_units}(C) and Fig.~\ref{fig:three_units}(C) still show 3 sections organized similar to the common functional units, but rougher in that the grouping is less crisp at the edges. In the /u/-/k/ motion, Fig.~\ref{fig:two_units}(C) and Fig.~\ref{fig:three_units}(C) again show that the division of functional units is less crisply segmented, compared with the functional units in Fig.~\ref{fig:two_units}(B) and Fig.~\ref{fig:three_units}(B). In Fig.~\ref{fig:two_units}(A) and Fig.~\ref{fig:three_units}(A) and Fig.~\ref{fig:two_units}(B) and Fig.~\ref{fig:three_units}(B), there is likely a compression/shortening of the tip-to-root region (green), compressing the tongue anteriorly-to-posteriorly, which elevates the posterior surface (red) up toward the velum. This compression (green) would reﬂect the line of action of the inferior longitudinal (IL) muscle. The /u/-/k/ motion in Fig.~\ref{fig:two_units}(C) and Fig.~\ref{fig:three_units}(C) shows a single unit for the upper anterior surface (red) and tip and a single unit for the posterior and root region (green). This could reﬂect a single unit for GG posterior and GH, whose muscles are parallel in the anterior-to-posterior direction. The GG posterior shortening pulls the root forward, and elevates the upper tongue. The GH shortening elevates the entire tongue as a unit.

Fig.~\ref{fig:box} illustrates the comparisons of the sizes of the identified functional units across subjects. Of note, all the data exhibited a normal distribution (Anderson-Darling test, $p$$>$0.05). For the transition of /\textschwa/ to /s/, the standard deviations of the sizes of the identified functional units from the previous approach and our approach were 10.4$\%$ and 6.8$\%$ for the two units ($p$$<$0.05), respectively. For the three units, the standard deviations of the sizes of the identified functional units from the previous approach and our approach were 7.9$\%$ and 2.2$\%$ for unit 1 ($p$$<$0.05), 8.1$\%$ and 2.3$\%$ for unit 2 ($p$$<$0.05), and 8.8$\%$ and 3.3$\%$ for unit 3 ($p$$<$0.05), respectively. The results indicated that our approach yielded reduced variability of the sizes of the functional units in terms of standard variations and that our approach and the previous approach showed significant statistical difference for all the units.

For the transition of /s/ to /u/, the standard deviations of the sizes of the functional units from the previous approach and our approach were 6.7$\%$ and 4.4$\%$ for the two units ($p$$<$0.05), respectively (see Fig.~\ref{fig:box}). For the three units, the standard deviations of the sizes of the functional units from the previous approach and our approach were 6.9$\%$ and 3.3$\%$ for unit 1 ($p$$<$0.05), 5.0$\%$ and 3.6$\%$ for unit 2 ($p$=0.1), and 8.0$\%$ and 2.1$\%$ for unit 3 ($p$$<$0.05), respectively. The results indicated that our approach yielded reduced variability of the sizes of the functional units in terms of standard variations, while our approach and the previous approach showed significant statistical difference except for unit 2 from three functional units.~~~~

For the transition of /u/ to /k/, the standard deviations of the sizes of the functional units from the previous approach and our approach were 10.2$\%$ and 3.7$\%$ for the two units ($p$=0.45), respectively. For the three units, the standard deviations of the sizes of the functional units from the previous approach and our approach were 9.9$\%$ and 6.5$\%$ for unit 1 ($p$$<$0.05), 7.1$\%$ and 5.9$\%$ for unit 2 ($p$=0.23), and 9.3$\%$ and 6.7$\%$ for unit 3 ($p$=0.09), respectively. We note that the results in Fig.~\ref{fig:two_units}(C) and Fig.~\ref{fig:three_units}(C) were identified by the previous approach~\citep{woo2018sparse} in the atlas space, while the results in Fig.~\ref{fig:two_units}(B) and Fig.~\ref{fig:three_units}(B) were co-identified with the common functional units in Fig.~\ref{fig:two_units}(A) and Fig.~\ref{fig:three_units}(A). The results indicated that our approach yielded reduced variability of the sizes of the functional units in terms of standard variations and that our approach and the previous approach showed significant statistical difference except for units 2 and 3 from three functional units.

%% Table 1 %%%%%%%%%%%%%%%%%%%%%%%%%%%%%%%%%%%%%%%%%%%%%%%%%%%%%%%
%\begin{table*}[h]
%\caption{Statistics of the sizes of functional units (Mean$\pm$SD)}
%\centering
%\begin{tabular}{c||c|c|c|c|c|c} 
% \hline 
% \multirow{2}{*}{Methods} & \multicolumn{2}{|c|} {/\textschwa/-/s/} & \multicolumn{2}{|c|} {/s/-/u/} & \multicolumn{2}{|c} {/u/-/k/} \\
% \cline{2-7}
%  & Unit 1 & Unit 2 & Unit 1 & Unit 2 & Unit 1 & Unit 2 \\
%   \hline \hline
%  Previous & 51.8$\pm$10.6 & 48.2$\pm$10.6 & 55.4$\pm$7.7 & 44.6$\pm$7.7 & 48.2$\pm$10.2 & 51.8$\pm$10.2 \\ \hline
%  Ours & 52.1$\pm$9.8 & 47.9$\pm$9.8 & 42.9$\pm$7.5 & 57.1$\pm$7.5 & 47.1$\pm$6.3 & 52.9$\pm$6.3 \\
% \hline
%\end{tabular}\label{table:stat_2}
%\end{table*}
%%%%%%%%%%%%%%%%%

%% Table 1 %%%%%%%%%%%%%%%%%%%%%%%%%%%%%%%%%%%%%%%%%%%%%%%%%%%%%%%
%\begin{table*}[t]
%\caption{Statistics of the sizes of functional units (Mean$\pm$SD)}
%\centering
%\begin{tabular}{c||c|c|c|c|c|c|c|c|c} 
% \hline 
% \multirow{2}{*}{Methods} & \multicolumn{3}{|c|} {/\textschwa/-/s/} & \multicolumn{3}{|c|} {/s/-/u/} & \multicolumn{3}{|c} {/u/-/k/} \\
% \cline{2-10}
%  & Unit 1 & Unit 2 & Unit 3 & Unit 1 & Unit 2 & Unit 3 & Unit 1 & Unit 2 & Unit 3 \\
%   \hline \hline
%  Previous & 33.9$\pm$8.8 & 34.6$\pm$6.5 & 31.5$\pm$5.4 & 34.2$\pm$5.6 & 33.5$\pm$4.6 & 32.3$\pm$6.7 & 31.9$\pm$6.0 & 32.1$\pm$9.3 & 35.9$\pm$8.5 \\ \hline
%  Ours & 34.6$\pm$4.8 & 31.9$\pm$4.5 & 32.3$\pm$5.2 & 32.0$\pm$4.2 & 33.7$\pm$5.1 & 34.3$\pm$4.8 & 32.4$\pm$4.1 & 33.0$\pm$5.6 & 34.6$\pm$6.2 \\
% \hline
%\end{tabular}\label{table:stat_3}
%\end{table*}
%%%%%%%%%%%%%%%%%

\section{Discussion}\label{sec:discussion}

The quest for identifying intrinsic ``dimension-reduced modular structures''---i.e., functional units---has been central to research on speech production, including motor control theories from different perspectives. Early findings~\citep{ohman1967numerical, mermelstein1973articulatory} indicated that the tongue is separated into tip and body carrying out ``quasi-independent'' motions. A recent study~\citep{stone2004functional} suggested that the tongue could be further divided into the anterior, dorsal, middle, and posterior regions carrying out ``quasi-independent'' motions. Additionally, there is a great deal of work investigating factor analytic models, including Principal Component Analysis (PCA)~\citep{slud2002principal, stone1997principal, stone2014tongue, xing2016analysis} and NMF~\citep{ramanarayanan2013spatio, woo2018sparse}, to represent tongue motions as linear combinations of the basic factors. Our study furthers this underlying framework via a data-driven approach in which any different size, shape, and region of the tongue can constitute this modular structure according to the task at hand. This is made possible, in part, owing to recent technological advancements in MR imaging and analysis and machine learning that allow us to examine both tongue structure and function at an unprecedented resolution and accuracy.

The successful speech movement requires the orchestration of a highly flexible configuration of intrinsic and extrinsic muscles of the tongue and the vocal tract articulators. The cortical control of articulation is known to be carried out by the ventral sensorimotor cortex~\cite{bouchard2013functional}. The production of intelligible speech arises from a coordinated motor pattern by means of a set of primitive or modular representations~\citep{browman1992articulatory,galantucci2006motor}. To mine such a modular structure in the tongue inherent in speech movements using NMF, \cite{woo2018sparse} proposed to incorporate two additional constraints, including sparsity and manifold geometry about the motion patterns, to determine a set of optimized and geometrically meaningful structures. This graph-regularized sparse NMF formulation allows computing a low-dimensional yet interpretable subspace, followed by identifying subject-specific functional units via spectral clustering. More recently, \cite{woo2020identifying} investigated the use of the same sparse NMF framework in a groupwise setting to co-identify the common and subject-specific functional units to increase interpretability due to large variability in the identified functional units across subjects. In the present work, we further proposed a joint deep graph-regularized sparse NMF and spectral clustering to co-identify the common and subject-specific functional units. This, in turn, increased interpretability and decreased size variability in the identified functional units compared with the previous approach \citep{woo2018sparse}. In addition, the identified subject-specific functional units are jointly obtained alongside the common functional units, thereby greatly facilitating the comparison of each subject with another.

To achieve deep NMF, we converted the standard NMF with sparse and graph regularizations into modular architectures using unfolding ISTA to learn building blocks and associated weighting map. The deep NMF using unfolding ISTA~\citep{gregor2010learning} has been studied previously, but it is worth noting that, to our knowledge, this is the first attempt at incorporating both sparse and graph regularizations into the ISTA framework. In addition, we further introduced a common low-dimensional subspace that can learn the common weighting map jointly with subject-specific weighting maps across subjects. 

The use of a deep variant of NMF based on ISTA is uniquely important to decompose the complex muscle coordination patterns into modular components or functional units over other alternative statistical techniques of matrix factorization, such as shallow NMF, PCA, Independent Component Analysis (ICA), or factor analysis. First and foremost, shallow NMF and its variants have been widely used for research on muscle synergies due to their great interpretability~\citep{shourijeh2016approach,ting2005limited,torres2007muscle,bruton2018synergies}. The shallow model, however, learns functional units or synergies by directly mapping the internal tongue motion to its underlying subspace. Successful tongue movement hinges on the orchestration of a complex set of neural activations of numerous intrinsic and extrinsic tongue muscles.~As such, considering the complex tongue structure and function, it is highly likely that the mapping between the internal tongue motion and its underlying low-dimensional subspace contains rather complex hierarchical information, which may not be captured by shallow NMF-based approaches. Second, PCA has been the most widely used method applied to kinematic data~\citep{wang2013review}, while NMF-based approaches are the best suited to studying muscle synergies because of their ability to handle the non-negative nature of muscle activation signals~\citep{bruton2018synergies}. More specifically, while both NMF and PCA learn low-dimentional building blocks and their weighting map, the non-negative constraints imposed on the decomposition process in NMF lead to a marked difference between the two methods. For example, the obtained building blocks from NMF are independent. If the distribution of the data is Gaussian, by contrast, then the obtained building blocks from PCA are orthogonal and independent. Otherwise, these building blocks will merely be uncorrelated, not necessarily independent. Therefore, NMF-based approaches learn interpretable and parts-based representations in that a set of components is combined to form a whole in a non-subtractive manner. In contrast, PCA represents each data as a linear combination of a limited number of building blocks that explain the maximal amount of variance. Since there are no non-negative constraints in PCA, the linear combination may mix the elements of the building blocks and weighting map, which can cancel each other out. Consequently, there is a lack of physical meaningfulness of the building blocks.~\citep{woo2018sparse}

There are a few limitations in this work. First, quantitative evaluation of the proposed work in the context of \textit{in vivo} tongue motion is a challenging task. The notion of accuracy within our unsupervised learning setting is ill-posed, as accurate validation is impossible due to the lack of ground truth other than simulation studies and visual assessment with a thorough knowledge of tongue structure and function. In the present work, a tongue motion simulator based on a vocal tract atlas~\citep{woo2015high} was used to generate Lagrangian tongue motion. With this simulator alongside the ground truth, we were able to validate our method, showing superior performance over the comparison methods. Second, the ideal method to measure motor control is electromyography (EMG) because it records the activation of muscles. However, the muscles of the tongue are orthogonal and almost entirely interdigitated, making it almost impossible to disambiguate one direction of fiber activation from another. Moreover, unlike typical skeletal muscles, the muscles of the tongue are not designed to move a bone around a joint, but to deform its surface to shape the vocal tract tube. Only imaging methods, including MRI and ultrasound, can measure speech motor control, while not interfering with speech. Ultrasound has also been used to study tongue motion, but it only records the tongue surface, not internal motion, and cannot see the structures beyond the tongue surface as the sound does not penetrate beyond the first tissue interface. Therefore, tagged MRI is the only modality that can be used to study speech motor control to the best of our knowledge. Third, in this work, we chose the number of clusters without a principled approach. In addition, there were a few parameters that we tuned with the help of the 3D tongue simulator. The development of a new approach to determine the optimal number of clusters in conjunction with optimization parameters is a subject for future research. Finally, in our simulation studies, while we used representative simulation datasets to test our approach, the number of sample size is too small to compute statistical significance. In our future work, we will increase the number of datasets to compute statistical diffrence between different approaches.

There are a few ways to expand on this work. First, the human tongue consists of numerous intrinsic and extrinsic muscles, each of which has distinct roles to compress and expand tissue points. For example, GG has a muscular architecture that locally activates different parts of the muscle, from GG anterior to GG posterior~\citep{miyawaki1975preliminary,stone2004functional}. As such, identifying such fine-grained local functional units within a single muscle or a subset of muscles in a hierarchical manner would reveal new insights into the mechanisms of how different elements of muscular architecture interact with each other. In order to accurately localize the internal muscles, structural MRI or diffusion MRI is needed as they can provide the location of the internal muscles or fiber architecture, respectively. Accurate registration~\citep{woo2014multimodal} is needed to put the imaging data into correspondence for both controls and patients~\citep{liu2021symmetric}. Second, various intra-subject variabilities in speech articulation is not fully explored in this work. In our future work, we will investigate functional units of a range of motion patterns having intra-subject variability to see the central tendency and its variability in the identified functional units. Finally, our framework can be applied to patient populations, such as those with amyotrophic lateral sclerosis~\citep{xing2018strain, lee2018magnetic} or tongue cancer with speech or swallowing impairments~\citep{woo2019differentiating}; assessing how local functional units adapt after a variety of treatments can potentially advance therapeutic, rehabilitative, and surgical procedures. For example, our prior work~\citep{xing2019atlas} found an increased correlation between the floor-of-mouth muscle group and internal tongue muscle group for tongue cancer patients compared with healthy subjects to compensate for their post-surgery function loss.~Our functional units analysis will further shed light on how patients adapt their speech movements depending on different tumor size and location as well as treatment methods.

To the best of our knowledge, this is the first report identifying common and subject-specific functional units from cine and tagged MRI. The atlas constructed from cine MRI was used as a reference anatomical configuration for subsequent analyses to identify and visualize the functional units of the internal motion patterns during speech. In this way, it was possible to contrast and compare the identified functional units across subjects that were not biased by each subject's anatomical characteristics. In addition, the proposed work furthered this underlying concept in which constructing the atlas of functional units was carried out in a low-dimensional subspace, since correspondences across subjects in the low-dimensional subspace were guaranteed through the reference material coordinate system. Therefore, the proposed work holds promise to provide a link between internal tongue motion and underlying low-dimensional subspace, thereby advancing our understanding of the inner workings of the tongue during speech. In addition, the identified common and subject-specific functional units could offer a unique resource in the scientific research community and open new vistas for functional studies of the tongue.

\section{Conclusion}\label{sec:conclusion}

In this work, we presented a new method to jointly identify common and subject-specific functional units. To address the limitations of shallow NMF and identify comparable and interpretable functional units across subjects, a deep joint NMF framework incorporating sparse and graph regularizations was proposed. Our proposed method was extensively validated on synthetic and \textit{in vivo} tongue motion data to demonstrate the benefit of its novel features. Our results show that our method can determine the common and subject-specific functional units with increased interpretability and decreased size variability.

\section*{Acknowledgments}
This work is partially supported by NIH R01DC014717, R01DC018511, R01CA133015, R21DC016047, R00DC012575, P41EB022544 and NSF 1504804 PoLS. 

%%Harvard
\bibliographystyle{model1a-num-names}

\end{document}